\documentclass[10pt,twocolumn,letterpaper]{article}

\usepackage[pagenumbers]{cvpr} 
\usepackage{marvosym}
\usepackage[dvipsnames]{xcolor}

\usepackage{graphicx}
\usepackage{subcaption}
\usepackage{bm}
\usepackage{booktabs}
\usepackage{siunitx}
\usepackage{colortbl, color}
\usepackage{multirow}
\usepackage{tabularx}
\usepackage{pifont}

\usepackage{xcolor}
\definecolor{cvprblue}{rgb}{0.21,0.49,0.74}

\usepackage{pifont}
\usepackage{algorithm}
\usepackage{algorithmic}

\usepackage[pagebackref,breaklinks,colorlinks,citecolor=cvprblue]{hyperref}

\def\paperID{*****} 
\def\confName{CVPR}
\def\confYear{2025}

\title{Event-Based Eye Tracking. 2025 Event-based Vision Workshop}

\author{
Qinyu Chen$^{1 \raisebox{-0.6ex}{\textsuperscript{\Letter}}}$ \and
Chang Gao$^{2}$ \and
Min Liu$^{3}$ \and
Daniele Perrone$^{4}$ \and
Yan Ru Pei$^{5}$\and
Zuowen Wang$^{6}$\and
Zhuo Zou$^{7}$\and
Shihang Tan$^{7}$\and
Tao Han$^{2}$ \and
Guorui Lu$^{1}$ \and
Zhen Xu$^{1}$ \and
Junyuan Ding$^{3}$ \and
Ziteng Wang$^{3}$ \and
Zongwei Wu$^{8}$ \and
Han Han$^{9}$\and
Yuliang Wu$^{9}$\and
Jinze Chen$^{9}$\and
Wei Zhai$^{9}$ \and
Yang Cao$^{9}$ \and
Zheng-jun Zha$^{9}$ \and
Nuwan Bandara $^{10}$\and
Thivya Kandappu $^{10}$\and
Archan Misra $^{10}$\and
Xiaopeng Lin $^{11}$\and
Hongxiang Huang $^{11}$\and
Hongwei Ren $^{11}$\and
Bojun Cheng $^{11}$\and
Hoang M. Truong$^{12, 13}$ \and
Vinh-Thuan Ly$^{12, 13}$ \and
Huy G. Tran$^{12, 13}$ \and
Thuan-Phat Nguyen$^{12, 13}$ \and
Tram T. Doan$^{12, 13}$ \\
\normalsize $^{1}$ Leiden University
\quad \normalsize $^{2}$Delft University of Technology
\quad \normalsize $^{3}$DVSense
\quad \normalsize $^{4}$Prophesee
\quad \normalsize$^{5}$ NVIDIA 
\\
\normalsize$^{6}$ Institute of Neuroinformatics, UZH/ETH Zurich
\quad \normalsize $^{7}$ Fudan University
\quad \normalsize$^{8}$ University of W\"urzburg 
\\ 
\normalsize$^{9}$ University of Science and Technology of China
\quad \normalsize$^{10}$ Singapore Management University
\\ \normalsize$^{11}$ The Hong Kong University of Science and Technology (Guangzhou)
\\ \normalsize$^{12}$ University of Science, VNU-HCM, Ho Chi Minh City, Vietnam
\\ \normalsize$^{13}$ Vietnam National University, Ho Chi Minh City, Vietnam
}

\begin{document}

\maketitle

\let\thefootnote\relax\footnotetext{
$\raisebox{-1.0ex}{\large\textsuperscript{\Letter}}$ Qinyu Chen (q.chen@liacs.leidenuniv.nl) is the corresponding author.\\
Challenge website: \url{https://lab-ics.github.io/3et-2025.github.io/}. Demonstration code repository: \url{https://github.com/EETChallenge/3et_challenge_2025}. Challenge Kaggle website: \url{https://www.kaggle.com/competitions/event-based-eye-tracking-cvpr-2025/overview}. 3ET+ Dataset: \url{https://www.kaggle.com/competitions/event-based-eye-tracking-cvpr-2025/data}. Event-based Vision workshop 2025 host website: \url{https://tub-rip.github.io/eventvision2025/}. 
} 

\begin{abstract}
This survey serves as a review for the 2025 Event-Based Eye Tracking Challenge organized as part of the 2025 CVPR event-based vision workshop. This challenge focuses on the task of predicting the pupil center by processing event camera recorded eye movement. We review and summarize the innovative methods from teams rank the top in the challenge to advance future event-based eye tracking research. In each method, accuracy, model size, and number of operations are reported. In this survey, we also discuss event-based eye tracking from the perspective of hardware design.
\end{abstract}

\setlength{\abovedisplayskip}{1pt}
\setlength{\belowdisplayskip}{1pt}

\section{\textcolor{black}{Introduction}}

With the rapid evolution of augmented and virtual reality (AR/VR) technologies, advanced by the consumer electronics industry particularly, the role of accurate and responsive eye-tracking systems has become increasingly important. For instance, the Apple Vision Pro features an advanced eye-tracking system that uses high-speed infrared cameras and LED illuminators to monitor eye movements with remarkable precision. This technology aims for accurate gaze estimation, allowing users to interact intuitively by simply looking at objects and confirming selections with subtle hand gestures.
Beyond human-computer interaction applications, eye-tracking technology is also emerging as a valuable tool in the domain of healthcare. Tasks such as gaze estimation, pupil shape tracking, and eye movement analysis offer powerful, non-invasive methods for detecting and monitoring neurological disorders, including Parkinson’s and Alzheimer’s diseases~\cite{pretegiani2017eye, duan2019dataset, Lee2000Schizophrenia}.

\begin{figure*}
    \centering
    \includegraphics[width=1.0\linewidth]{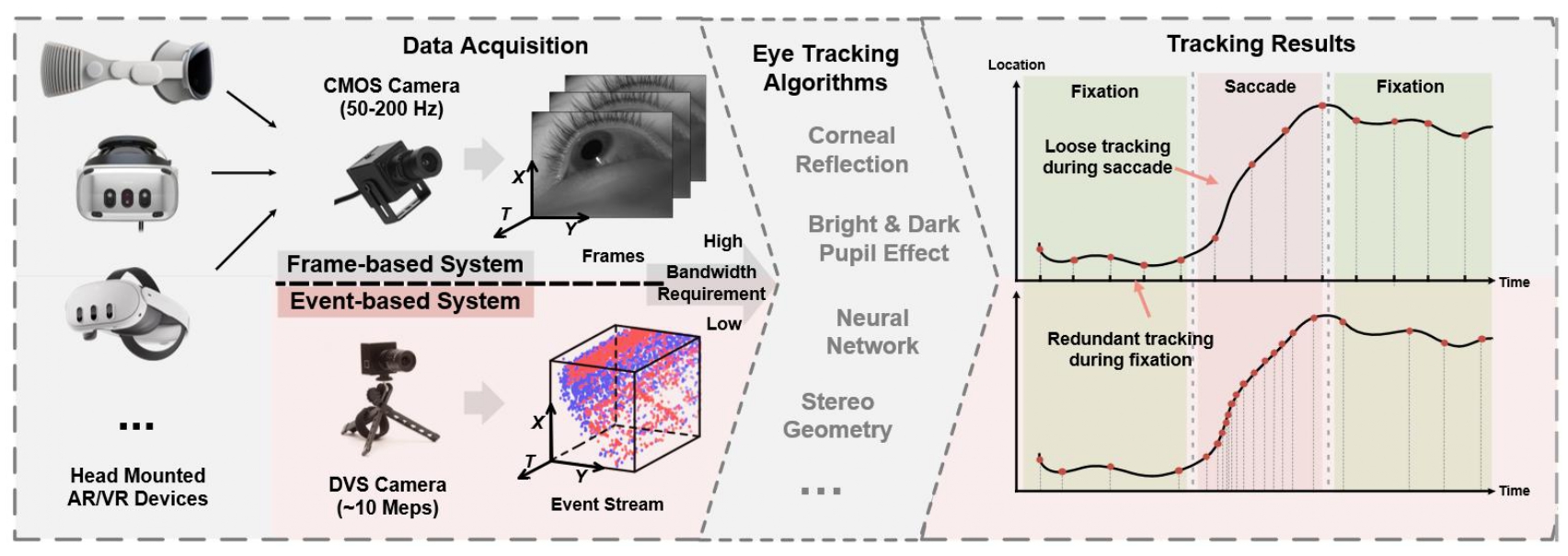}
    \caption{Comparison of the processing flow and estimation patterns between frame-based and event-based systems for eye tracking. Adapted from \cite{tan2025etprocessor}.}
    \label{fig:intro_flow}
\end{figure*}

Mobile platforms are typically constrained by strict power and computing budgets, making it challenging to deploy complex software and algorithms. In addition, eye-tracking tasks demand high-frequency sensory sampling, which imposes additional burdens on hardware and data pipelines. 
For mobile AR/VR applications, an eye-tracking system should be lightweight to integrate seamlessly into head-mounted devices while supporting high temporal resolution and good task accuracy. This is particularly critical considering that the human eye can move at angular velocities exceeding 300°/s and accelerations up to 24,000°/s$^2$~\cite{10000hz2021event}, necessitating sampling rates in the kilohertz range to accurately capture the onset and dynamics of fast eye movements. However, achieving such a high frame rate is challenging for wearable devices, which must operate at low power levels, typically in the milliwatt range. Most head-mounted devices rely on frame-based eye-tracking systems. a recent study~\cite{stein2021comparison} indicates that many such systems experience tracking delays ranging from 45 to 81\,ms, which is insufficient for capturing rapid eye movements that require kilohertz-level frame rates. Furthermore, frame-based sensors capable of operating at kilohertz tend to consume a large amount of power. The resulting high data throughput also demands considerable bandwidth and energy for transmission and computation, making real-time deployment on low-power wearable platforms difficult.

Event cameras, also named Dynamic Vision Sensors (DVS)~\cite{dvs1, dvs2, dvs3}, are unique vision sensors that offer several potential advantages for eye-tracking in mobile devices. Different from traditional cameras, event cameras asynchronously detect log intensity changes in brightness that exceed a certain threshold. This unique way of sensing induces spatiotemporally sparse camera outputs (events). Many research works~\cite{dvs_transformer2022Wang, Messikommer20eccv, Sabater_2022_CVPR, pointnet_event_camera} have been proposed to exploit this spatiotemporal sparsity, aiming to reduce the hardware platform requirements of computation and energy. In addition, the temporal precision of eye-tracking tasks could benefit from the high temporal resolution data property of event cameras. 

Fig.~\ref{fig:intro_flow} shows some of the most commonly used head-mounted devices and their corresponding frame-based eye-tracking processing flow and systems in comparison to the event-based solution utilizing a DVS. The event-based approach shows promising potential in providing more robust tracking performance while requiring small bandwidth and less power consumption.
These unique characteristics make event cameras highly suitable for high-speed, low-power eye tracking: they produce less data and reduce processing needs during fixation while still capturing fast and subtle eye movements during saccades. Previous event-based eye-tracking studies have shown promising results~\cite{3et, bonazzi2024retina, EGaze2024li, 10000hz2021event, pei2024spatiotemporal, zhao2024ev, li2023track,wang2024mambapupil,stoffregen2022event,lin2024fapnet,ding2024facet}.

The 2025 Event-Based Eye Tracking Challenge is set to explore algorithmic potentials for event-based eye-tracking. By emphasizing efficient methods that can extract eye-position-relevant information from sparse event streams, the challenge seeks to drive advancements in event-based eye-tracking that are fast and efficient and are suitable for wearable healthcare devices and real-time AR/VR applications.

\section{\textcolor{black}{Event-based Eye Tracking Challenge}}

\subsection{\textcolor{black}{Introduction of the \text{3ET+} dataset}}
The \texttt{3ET+} dataset~\cite{wang2024ais_event, chen2025eventvision_event} offers a comprehensive benchmark for event-based eye-tracking research. Captured using a DVXplorer Mini~\cite{xplorer} event camera, it features recordings from 13 participants, each contributing between 2 to 6 sessions. During each session, subjects are required to perform five distinct eye movement tasks: \textit{random movements}, \textit{saccades}, \textit{text reading}, \textit{smooth pursuits}, and \textit{blinks}. Ground truth annotations were provided at 100 Hz, including (1) a binary label indicating the presence or absence of a blink and (2) manually labeled pupil center spatial coordinates for precise tracking.

\subsection{\textcolor{black}{Task description}} \label{sec:main_task_description}
\begin{itemize}
    \item \textbf{Input:} Raw events $(x_i, y_i, t_i, p_i)$ from recording eye movements. $(x_i, y_i)$ are spatial coordinate, $t_i$ is the timestamp and $p_i$ is the polarity of the event $e_i$.
    \item \textbf{Task:}  Predict the spatial coordinates $x_i, y_i$ of the pupil center at the specified timestamps, matching the frequency of the ground truth, within the input space.
    \item \textbf{Metric:}
The evaluation metric in this year’s challenge differs from that of last year \cite{wang2024ais_event}. This year, the primary metric on the Kaggle leaderboard is pixel error, defined as the Euclidean distance between the spatial coordinates of the predicted label and the ground truth.  
In contrast, last year’s leaderboard used p-accuracy as the main evaluation metric. Under this metric, a prediction is considered correct if the pixel error is within p pixels. The leaderboard used a threshold of p = 10 pixels.
The shift from p-accuracy to pixel error was motivated by the observation that last year’s models often achieved near-perfect p\_10 scores (close to 100\%), making it difficult to distinguish between high-performing models and limiting the potential for further improvement.
\end{itemize}

\subsection{\textcolor{black}{Data loading and training pipeline}}
The challenge provided participants with a convenient data loading and training pipeline. The data loader was designed to be compatible with the Tonic library~\cite{tonic_lenz_gregor}, allowing users to experiment with different event-based feature representations.
It also supports caching of generated features either in memory or on disk during the first training epoch, enabling faster data loading in subsequent epochs. 
For the training process, participants could easily integrate their own deep learning models and adjust hyperparameters as needed. A machine learning monitoring library, namely the MLFlow library~\cite{mlflow}, was also provided in the challenge pipeline code for the participants to monitor various metrics and to record the hyperparameters, as well as the checkpoints.

\subsection{Challenge phases}

The challenge is organized into three key phases:

\begin{enumerate}
    \item \textbf{Preparation Phase} (Before 15. Feb. 2025): Finalize the challenge dataset, develop the code pipeline, set up the competition website, and prepare the Kaggle platform.
    \item \textbf{Competition Phase} (Starting 15. Feb.  2025): The Kaggle competition officially launches. Teams can register, download the dataset, and begin working on their solutions.
    \item \textbf{Submission and Evaluation Phase} (15. Mar. 2025): The submission portal closes. Private leaderboard scores are revealed, and top-performing teams are invited to submit their factsheets and source code. Selected teams are also encouraged to contribute a paper detailing their methods to the associated workshop.
\end{enumerate}

\subsection{Related Challenge} 
This challenge is part of the 2025 Event-based Vision Workshop and serves as the second edition of the Event-based Eye Tracking Challenge, following the first iteration presented in~\cite{wang2024ais_event}. The main improvements in this year’s challenge include upgrading the label frequency from 20 Hz to 100 Hz for finer temporal resolution, and changing the evaluation metric from p-accuracy to pixel error for more precise performance differentiation. The results demonstrate a notable improvement in tracking accuracy: four participating teams achieved a pixel error below 1.7, outperforming the best results from the previous year.

\section{\textcolor{black}{Challenge Results}}
We summarize the main evaluation results from the participating teams in Tab.~\ref{tab:results}. The pixel error described in Sec.~\ref{sec:main_task_description} is used as the primary evaluation metric for the Kaggle competition ranking. The model size is reported in Tab.~\ref{tab:results} and the number of operations is reported in the sections of each team.

\begin{table}[ht]
\centering
\small
\setlength{\tabcolsep}{5pt}
\begin{tabular}{lc|cc}
\toprule
\textbf{Team} & \textbf{Rank}  & \textbf{Pixel error} & \textbf{Param (M)} \\
\midrule
USTCEventGroup & 1&  1.14 & 7.1   \\
EyeTracking@SMU & 2& 1.42 &  0.8  \\
HKUSTGZ & 3 &1.50 & 3.0   \\
CherryChums & 4& 1.61 & 0.8   \\
\bottomrule
\end{tabular}
\caption{Final results from the top performing teams. Details can be found in this survey paper.}
\label{tab:results}
\end{table}

\subsection{Architectures and main ideas}
The methods proposed by the participating teams range from novel pre-processing techniques and custom model architecture designs to motion-aware postprocessing method. The major novelties and contributions are summarized as follows:

\textbf{Modeling Short- and Long-Term Temporal Dependencies} 
Effectively capturing both short- and long-term temporal dynamics is important for accurate event-based eye tracking. Team USTCEventGroup modeled short-term motion with Bi-GRU and, in the following, long-term dependencies using a self-attention module enhanced with bidirectional relative positional attention. Team HKUSTGZ adopted a 3D CNN for capturing the implicit short-term temporal dynamics of the eye
movements, while long-term dependencies were handled by a cascade of GRU and Mamba modules.

\textbf{Data Augmentation and Generalization Strategies} 
Team CherryChums implemented a practical augmentation pipeline, including temporal shift, spatial flip, and random event deletion, simulating real-world perturbations such as motion jitter, mirroring, and sensor dropout. In addition, pretraining on an external event dataset (synthetic 3ET dataset~\cite{3et}) was used by Team USTCEventGroup to improve generalization and provide stronger initialization under limited training data conditions.

\textbf{Model-Agnostic Inference-Time Post-processing} 
Blink artifacts can interrupt event streams and cause erroneous gaze predictions, and temporal inconsistency may lead to unstable, non-smooth gaze trajectories.
Team EyeTracking@SMU proposed two lightweight post-processing techniques: 1) Motion-Aware Median Filtering (M2F) to ensure temporal smoothness by adaptively smoothing gaze trajectories based on motion variance. 2) Optical Flow-based Refinement (OFE) to adjusts predictions using local event motion flow to correct spatial misalignments.
These steps require no retraining or model changes and can be flexibly applied to any existing model.

\subsection{Participants}
There were, in total, 22 user accounts registered and participated the challenge, and 4 teams with private pixel error lower than 1.7 submitted factsheets describing their methods. 

\subsection{Inclusiveness and fairness}
To ensure inclusiveness and fairness, several initiatives were implemented during the challenge. First, the dataset and task design were carefully optimized to reduce computational demands, making participation accessible to teams with limited hardware resources. Second, a ready-to-use training and testing pipeline was provided, allowing even those with minimal experience in event-based data to quickly get started. Finally, submitting source code alongside the factsheets was mandated to guarantee the reproducibility of all results.

    

\section{Conclusion and Outlook}
Over the last two years, this series of challenges has significantly advanced the event-based eye tracking field. The participating teams demonstrated remarkable innovation through various approaches. Although teams reported basic metrics such as parameter counts and pixel error, a deeper understanding of model efficiency and computational costs remains essential to give more useful insights for hardware designers for edge AI hardware accelerators for AR/VR wearables and more. For example, we could explore metrics that capture the actual computational workload of models, such as arithmetic operations, the memory footprint of activations (feature maps)~\cite{Cai2020}, and analyze the level of sparsity if the neural network is optimized to induce spatial~\cite{Han2015,han2016eie,han2017ese,chen2020efficient}, temporal~\cite{liu2022,gao2020edgedrnn,Chen2025deltakws} or spatio-temporal sparsity~\cite{chen2022skydiver, gao2022spartus,hunter2022two,Liu2024} using tools like NeuroBench~\cite{yik2025neurobench}. Further discussion on hardware design can be seen in the Sec.~\ref{sec:discussion}.

\section*{Acknowledgements}
This challenge was partially (e.g., dataset collection) funded by the Swiss National Science Foundation and Innosuisse BRIDGE - Proof of Concept Project (40B1-0\_213731) and the NWO (Dutch Research Council) Talent Programme Veni AES 2023 (File number 21132). The dataset collection was partially supported by the 2023 Telluride Neuromorphic Cognition Engineering Workshop. This challenge was partially supported by DVsense.

\section{Challenge Teams and Methods}
\label{sec:teams}

The following sections present an overview of the top-performing solutions from the challenge. Each method description was authored and submitted by the respective teams as part of their contribution to this survey.


\subsection{Team: USTCEventGroup}


\begin{center}

\noindent\emph{Han Han, Yuliang Wu, Jinze Chen, Wei Zhai, Yang Cao, Zheng-jun Zha}

\noindent\emph{University of Science and Technology of China}

\noindent{\emph{Contact: \url{hanh@mail.ustc.edu.cn}}}

\end{center}

\begin{figure*}[htbp!]
    \centering
\includegraphics[width=0.8\linewidth]{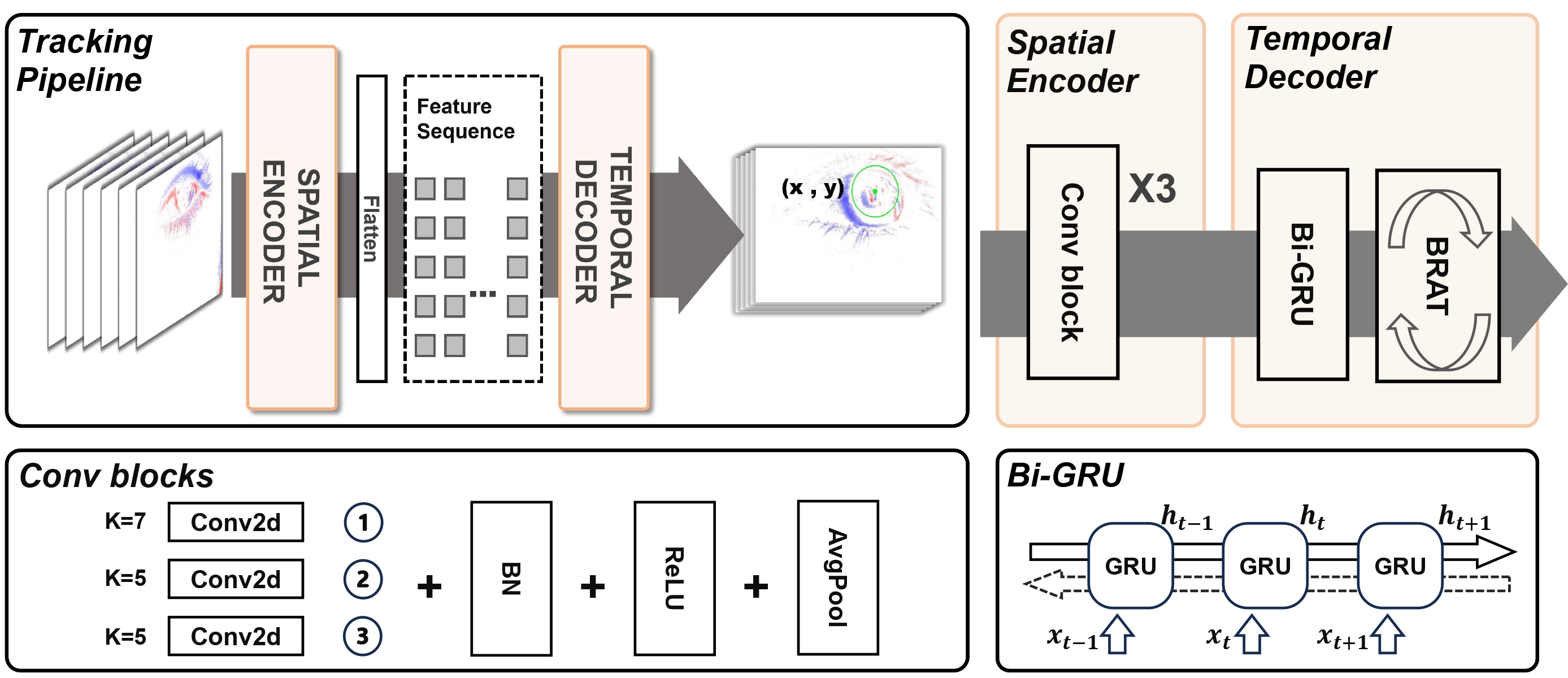}
    \caption{BRAT network by Team USTCEventGroup.}
    \label{fig:BRAT_Main}
\end{figure*}

\paragraph{Description.} 
The USTCEventGroup proposed the Bidirectional Relative Positional Attention Transformer (BRAT) method as shown in \cref{fig:BRAT_Main}. 
This network is composed of a spatial encoder and a temporal decoder. The former utilizes a CNN to extract geometric structural features from event representations, while the latter combines a Bi-GRU block and the BRAT block to analyze temporal motion patterns and accurately localize pupil positions. 
In the spatial feature extraction phase, the input binary representation is initially processed by a convolutional layer with a kernel size of 3, resulting in 32 channels. Subsequently, spatial features are extracted through three additional convolutional layers that use larger kernel sizes of 7, 5, and 5. After the first two large-kernel convolutional layer, pooling layers are applied. And following the third convolutional layer, the feature maps undergo a spatial dropout layer to mitigate overfitting.

\begin{figure*}[t]
    \centering
\includegraphics[width=0.9\linewidth]{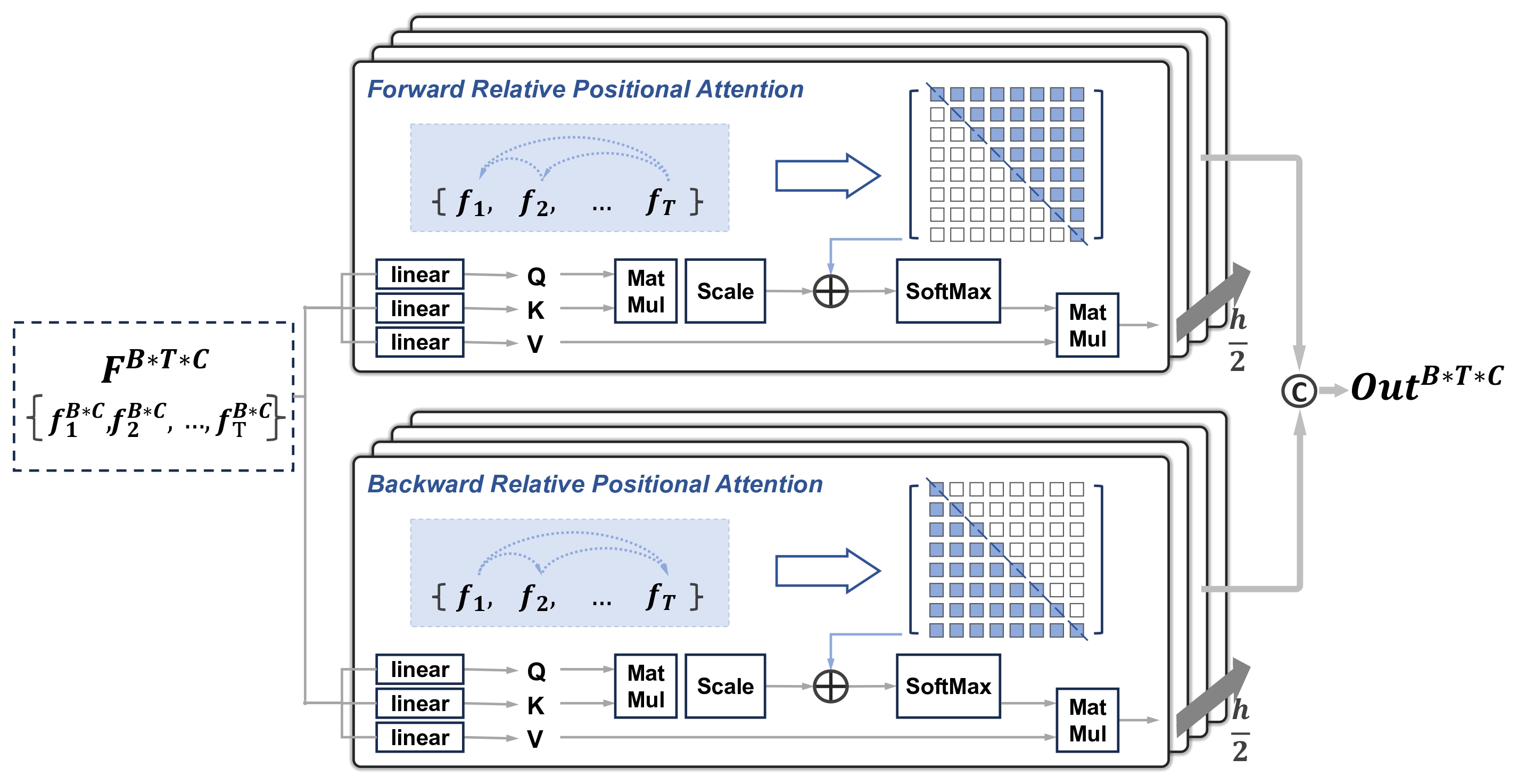}
    \caption{Bidirectional Relative Positional Attention.}
    \label{fig:BRA}
\end{figure*}

After extracting spatial information, the features are fed into bidirectional GRU blocks to capture short-term temporal patterns. They are then further processed within the BRAT architecture to model long-range dependencies. 
In BRAT architecture, the standard multi-head self-attention in transformer is replaced with a bidirectional variant that explicitly encodes relative temporal distances, see \cref{fig:BRA}. Given a sequence of length $T$ and $h$ attention heads, the output of each head is computed as:
\begin{align}
    \textbf{Attention}^{i} = \text{softmax}\left(\frac{\mathbf{Q}^{i} \mathbf{K}^{i\top}}{\sqrt{d_k}} + \mathbf{B}^{i}\right) \mathbf{V}^{i},
\end{align}
where $\mathbf{B}^{i} \in \mathbb{R}^{T \times T}$ is a relative position bias matrix designed to modulate attention weights based on temporal distance. It is decomposed into forward and backward components as:
\begin{align}
    \mathbf{B}^{i}_{forward} =
        \begin{cases}
        m^{i} \cdot (t - s), & t\ge s \\
        0, & t < s \\
        \end{cases},
\end{align}
\begin{align}
    \mathbf{B}^{i}_{backward} =
        \begin{cases}
        0, & t\ge s \\
        m^{i} \cdot (s - t), & t < s \\
        \end{cases},
\end{align}
where $m^{i}$ denotes the sensitivity of head $i$ to the relative position, generated via a monotonically decreasing linear mapping to progressively diminish attention to distant steps. 

Furthermore, to improve the robustness of the model the USTCEventGroup adopted a multi-time-step data sampling strategy during training. Specifically, a sliding window with a fixed length was applied over the long event sequence with a stride of 1, generating dense training samples. Within each window, frames were sampled at uniform intervals defined by a step size, allowing the model to observe motion over longer temporal spans while controlling input density. 
For training supervision, the loss is calculated by finding the squared differences between predicted values and true labels at each time step, then taking the square root and averaging over the time dimension:
\begin{align} 
Loss = \frac{1}{T}\sqrt{\sum_{t=1}^{T} (y_{t,pred} - y_{t,label})^2}. 
\end{align} 
This formulation captures the overall error across the temporal sequence, normalizing by the sequence length to account for varying durations.

\begin{table}[ht]
    \centering
    \begin{tabular}{cccc}
        \toprule
        \multicolumn{2}{c}{\textbf{Metric}} & \multicolumn{2}{c}{\textbf{Value}}\\
        \midrule
        \multicolumn{2}{c}{Param} & \multicolumn{2}{c}{7.1~M}\\
        \midrule
        \multicolumn{2}{c}{Number of MACs}& \multicolumn{2}{c}{2.9~G}\\
        \bottomrule
    \end{tabular}
    \caption{Model complexity of BRAT.}
    \label{tab:ustceventgroup:model scale}

\end{table}
\paragraph{Implementation Details.}
All experiments were conducted using PyTorch, employing Cosine Annealing Warm Restart as the learning rate scheduler, starting with an initial rate of 0.001. 
The model was initially pretrained on the 3ET simulation dataset~\cite{3et}, followed by training and evaluation on the 3ET+ dataset, which took approximately 24 hours for 800 epochs at a batch size of 32 on a single RTX 2080Ti GPU. 
\cref{tab:ustceventgroup:model scale} shows the model complexity metrics, and the model does not require additional stages for deployment.

\paragraph{Results.}

The BRAT proposed by USTCEventGroup achieved first place on the leaderboard. The results are in the Tab.~\ref{tab:ustceventgroup:valid_results}. The visualization results demonstrate the model's high accuracy and robust tracking capabilities in different cases. Visualization results and code are available in \url{https://github.com/hh-xiaohu/Event-based-Eye-Tracking-Challenge-Solution}.

\begin{table}[ht]
    \centering
    \begin{tabular}{cccc}
        \toprule
        \textbf{ p\_5} & \textbf{p\_10} & \textbf{p\_15} & \textbf{pixel error} \\
        \midrule
       
         0.978 & 0.995 & 1.000 & 1.14 \\
    
        \bottomrule
    \end{tabular}
    \caption{Validation results of BRAT.}
    \label{tab:ustceventgroup:valid_results}
\end{table}

\newcommand{\cmark}{\ding{51}}%
\newcommand{\xmark}{\ding{55}}%

\subsection{Team: EyeTracking$@$SMU}


\begin{center}

\noindent\emph{Nuwan Bandara, Thivya Kandappu, Archan Misra}

\noindent\emph{Singapore Management University}

\noindent{\emph{Contact: \url{thivyak@smu.edu.sg}}}

\end{center}

\paragraph{Description: Motivation \& Background. } 

In this work~\cite{bandara2025model}, our team specifically addresses several key limitations in the existing event-based eye tracking models. The first limitation is the handling of blink artifacts, which cause interruptions in the event data and lead to erroneous gaze predictions~\cite{zhang2024swift}. Another limitation is the temporal inconsistency often observed in the predictions, as eye movements are physiologically continuous and models sometimes fail to enforce this temporal smoothness, leading to abrupt gaze shifts that undermine tracking stability~\cite{holmqvist2011eye,bandara2024eyegraph}. Additionally, existing models often fail to fully leverage local event distributions, resulting in misaligned gaze predictions. These challenges, coupled with the inherent label sparsity of event datasets, make it difficult to develop a universally robust event-based tracking system.

To address these challenges, we propose a model-agnostic inference-time post-processing to enhance the accuracy and robustness of event-based eye tracking. Our approach targets the shortcomings of existing spatio-temporal models by introducing lightweight, post-processing techniques that can be integrated with any model without requiring retraining or architectural changes. This makes our method flexible and easily applicable to a wide range of existing models. The post-processing framework consists of two key components: (i) motion-aware median filtering (M2F), which enforces temporal smoothness by taking advantage of the continuous nature of eye movements, and (ii) optical flow-based local refinement (OFE), which improves spatial consistency by aligning gaze predictions with dominant motion patterns in the local event neighborhood. These refinements not only mitigate blinking artifacts but also ensure that gaze predictions remain temporally continuous and spatially accurate, even in the presence of rapid eye movements or motion artifacts.
\setlength{\textfloatsep}{0pt}
\begin{algorithm}[t]
\caption{Motion-aware median filtering}\label{algo:median}
\begin{algorithmic}[1]
\REQUIRE Original predictions $\{x_{pred}, y_{pred}\}$, base window for local motion variance estimation $w_{base}$, minimum allowed smoothing window $w_{min}$, maximum allowed smoothing window $w_{max}$, percentile to determine adaptive window size $p$, method $f(.)$ $\in$ \{displacement, velocity, acceleration, covariance, frequency\}
\STATE Output: filtered predictions $\{x_{(f, pred)}, y_{(f, pred)}\}$ 
\STATE local motion variance $\longleftarrow$ $f(\{x_{pred}, y_{pred}\}, w_{base})$
\STATE smoothened variance $\longleftarrow$ rolling mean($w_{base}$, local motion variance)
\STATE median window $\longleftarrow$ clipping(smoothened variance, $w_{min}$, $w_{max}$)
\STATE adaptive windows $\longleftarrow$ clipping($w_{min}$, $w_{max}$, rolling(median window, $w_{base}$, $p$))
\STATE $\{x_{(f, pred)}, y_{(f, pred)}\}$ $\longleftarrow$ rolling median($\{x_{pred}, y_{pred}\}$, adaptive windows) 
\end{algorithmic}
\end{algorithm}
\paragraph{Description: Inference-time Post-processing.} 

As discussed above, to address the shortcomings of existing methods in the inference stage, we propose to add two light-weight post-processing techniques specifically targeting the following limitations: (1) motion-aware median filtering (algorithm~\ref{algo:median}) to (a) ensure the temporal consistency of the predictions since the eye movements are physiologically bound to be continuous in spatial domain~\cite{holmqvist2011eye} and (b) reduce the blinking artifacts and (2) optical flow estimation in the local spatial neighbourhood (algorithm~\ref{algo:ofe}) to smoothly shift the original predictions if the flow vector at the original prediction is unaligned with the cumulative local neighbourhood flow direction. This (2) is specifically inspired by our empirical observations which hint that the original predictions tend to occupy a negligence towards the event motion flow in the local neighbourhood, suggesting a lack of attention to the local event distribution in the original models. 

\begin{algorithm}[!t]
\caption{Rule-based optical flow estimation for smooth shifts}\label{algo:ofe}
\begin{algorithmic}[1]
\REQUIRE Continuous event stream with $N$ number of events $E^v_{i, (t,x,y,p)}$ where $i \in \{1, N\}$, filtered predictions $\{x_{(f, pred)}, y_{(f, pred)}\}$, scaling parameter $\tau$, count threshold $c$, difference threshold $\gamma$
\STATE Output: Refined predictions $\{x_{(R, f, pred)}, y_{(R, f, pred)}\}$
\STATE timestep $\longleftarrow$ $\frac{E^v(i=N, t_{max}) - E^v(i=1, t_{min})}{|\{x_{(f, pred)}, y_{(f, pred)}\}|}$
\STATE previous timestamp $\longleftarrow$ $E^v(i=1, t_{min}) \in E^v_{i, (t,x,y,p)}$
\STATE ROI size $R$ $\longleftarrow$ $\tau \times 10$
\FOR{$j$, $(x^j_{(f, pred)}, y^j_{(f, pred)})$ $\in$ $\{x_{(f, pred)}, y_{(f, pred)}\}$}
    \STATE current timestamp $\longleftarrow$ previous timestamp $+$ $(j+1) \times timestep$
    \IF{$j > c$}
        \STATE difference in x $\longleftarrow$ absolute($x^j_{(f, pred)} - mean(\{x^{j-c:j}_{(f, pred)}\})$)
        \STATE difference in y $\longleftarrow$ absolute($y^j_{(f, pred)} - mean(\{y^{j-c:j}_{(f, pred)}\})$)
        \IF{difference in x $> \tau \times \gamma$ $\cup$ difference in y $> \tau \times \gamma$}
            \STATE $R$ $\longleftarrow$ $(1+c) \times \tau$
        \ELSE
            \STATE $R$ $\longleftarrow$ $(1-c) \times \tau$
        \ENDIF
    \ENDIF
    \STATE events in ROI $\longleftarrow$ $E^v(t\in$ \{previous timestamp, current timestamp\}, $x \in \{x^j_{(f, pred)}-R, x^j_{(f, pred)}+R\}$, $y \in \{y^j_{(f, pred)}-R, y^j_{(f, pred)}+R\}$ $) \in E^v_{i, (t,x,y,p)}$
    \STATE previous timestamp $\longleftarrow$ current timestamp
    \STATE $n$ $\longleftarrow$ |events in ROI|
    \IF{$ n > \tau \times 10$}
        \STATE $dx \longleftarrow 0; dy \longleftarrow 0$
        \FOR{$k \in \{1, n\}$}
            \STATE $dx +=$ events in ROI($x=k$) $-$ events in ROI($x=k-1$)
            \STATE $dy +=$ events in ROI($y=k$) $-$ events in ROI($y=k-1$)
            \IF{absolute($dx$) $> 0 \cup$ absolute($dy$) $> 0$}
                \STATE $x^j_{(R, f, pred)} \longleftarrow x^j_{(f, pred)} + \frac{dx}{\Vert dx, dy\Vert}$
                \STATE $y^j_{(R, f, pred)} \longleftarrow y^j_{(f, pred)} + \frac{dx}{\Vert dx, dy\Vert}$
            \ENDIF
        \ENDFOR
    \ENDIF
\ENDFOR
\end{algorithmic}
\end{algorithm}
\begin{table}[!t]
    \centering
    \begin{tabular}{c|c|c|c|c}
            \hline
        Method &  M2F & OFE & pixel error & pixel error \\
         &   &  & (Public) & (Private)\\
        \hline
        CG~\cite{3et} & \xmark & \xmark & 7.914 & 7.922\\
        CG~\cite{3et} & \cmark & \cmark & 7.494 & 7.504\\
        BB~\cite{pei2024spatiotemporal} & \xmark & \xmark & 1.431 & 1.500 \\
        BB~\cite{pei2024spatiotemporal} & \cmark & \xmark & 1.408 & 1.466 \\
        BB~\cite{pei2024spatiotemporal} & \cmark & \cmark & \textbf{1.382} & \textbf{1.423}\\
        \hline
    \end{tabular}
    \caption{Evaluation Results of Team EyeTracking$@$SMU's approach.}
    \label{tab:SMU_table}
\end{table}

\begin{table}[!t]
    \centering
    \begin{tabular}{c|c|c|c|c}
        \hline      
        Method &  M2F & OFE & Model size& $\#$ MACs\\
         &   &  & size&  \\
        \hline
        CG~\cite{3et} & \xmark & \xmark &  417K & 2716.419840M \\
        CG~\cite{3et} & \cmark & \cmark &  417K & 2716.420096M \\
        BB~\cite{pei2024spatiotemporal} & \xmark & \xmark & 809K &  59.537664M \\
        BB~\cite{pei2024spatiotemporal} & \cmark & \cmark & 809K & 59.537920M \\
        \hline
    \end{tabular}
    \caption{Computational complexity details of Team EyeTracking$@$SMU's approach.}
    \label{tab:SMU_table_macs}
\end{table}

More descriptively, in motion-aware filtering as shown in Algorithm~\ref{algo:median}, we first estimate the local motion variance in temporal dimension (i.e., within a set time window) using a set of alternative methods including $0^{th}$ to $2^{nd}$ order kinetics, covariance and frequency and subsequently, assign a median-based adaptive filter windows for each set time windows such that the kernel size for median filtering is adaptive and appropriate to the background pupil movement while also ensuring the temporal consistency. In contrast, in optical flow estimation as shown in Algorithm~\ref{algo:ofe}, we first estimate the appropriate size for the region of interest (ROI) around the filtered prediction using the first order derivatives of $x$ and $y$ and then, if the number of events within the selected ROI exceeds a set threshold, we accumulate and determine the cumulative vector trajectory of the events within ROI to softly shift the filtered prediction to further refine its spatial position.

\paragraph{Implementation Details: Base Models.} Since our proposed method is presented as a post-processing step and works in a model-agnostic fashion, we select two recent models as base models: CNN-GRU (CG)~\cite{3et}, and bigBrains (BB)~\cite{pei2024spatiotemporal}, to show the impact of the proposed pipeline towards improved pupil coordinate predictions in each case. To be specific, CG is a simple convolutional gated recurrent unit architecture which specifically designed for efficient spatio-temporal modelling to predict pupil coordinates from sparse event frames, whereas BB attempts to preserve causality and learn spatial relationships using a lightweight model consisting of spatial and temporal convolutions.
\begin{figure*}[!t]
    \centering
    \includegraphics[width=0.9\linewidth]{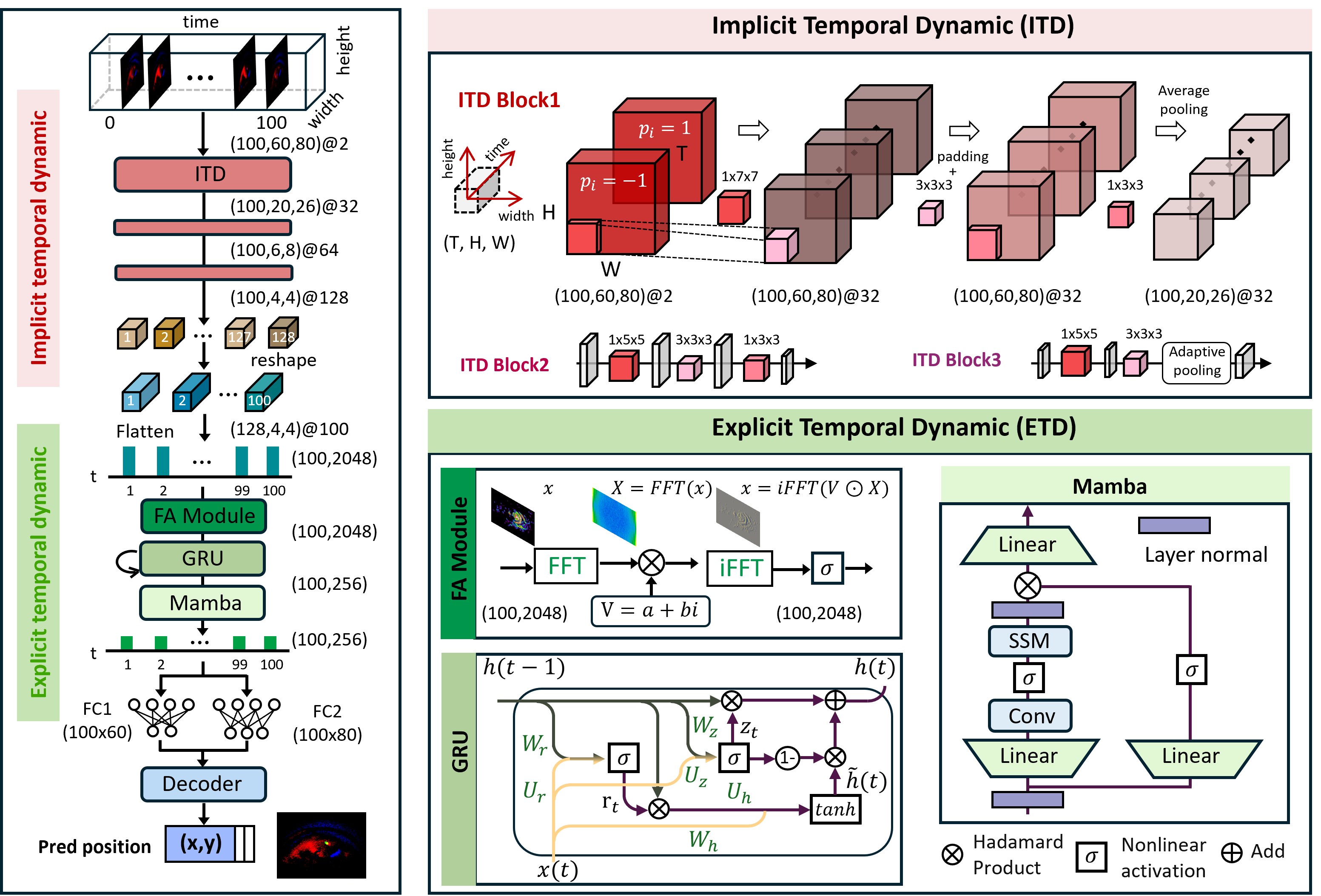}
    \caption{The architecture of TDTracker. TDTracker primarily comprises two components, Implicit Temporal Dynamic (ITD) and  Explicit Temporal Dynamic (ETD), with a structure featuring three ITD components to ensure effective feature abstraction. It employs a cascaded architecture of three distinct time series models to capture temporal information comprehensively.}
    \label{fig: TDTracker}
\end{figure*}
\paragraph{Results.} As shown in Tab.~\ref{tab:SMU_table}, both of our post processing techniques consistently improved the results of vanilla predictions of each method and thereby suggest the efficacy of the proposed model-agnostic post-processing methods. In addition, since our methods are executed at inference time as light-weight post-processing steps, we estimate the FLOPs of M2F and OFE to be $\approx 172$ and $\approx 340$ per prediction instance (i.e., per event frame), respectively, whereas the learnable parameter space is effectively null. As shown in Tab.~\ref{tab:SMU_table_macs}, we present in detail that our post-processing steps only add a negligible overhead to the base models in terms of computational complexity, despite consistently improving the vanilla prediction results of the base models. Our code is available at \href{https://github.com/eye-tracking-for-physiological-sensing/EyeLoRiN}{github/EyeLoRiN}.

\subsection{Team: HKUSTGZ}

\begin{center}

\noindent\emph{Xiaopeng Lin, Hongxiang Huang, Hongwei Ren, Yue Zhou, Bojun Cheng}

\noindent\emph{The Hong Kong University of Science and Technology (Guangzhou)}

\noindent{\emph{Contact: \url{bocheng@hkust-gz.edu.cn}}}
\end{center}

\paragraph{Description.} 
The HKUSTGZ team proposes the TDTracker framework \cite{ren2025exploring}, which is designed to address the challenges of high-speed, high-precision eye tracking using event-based cameras as shown in \cref{fig: TDTracker}. It consists of two main components: a 3D convolutional neural network (CNN) and a cascaded structure that includes a Frequency-Aware Module \cite{ren2024frequency}, Gated Recurrent Units (GRU) \cite{gru2014}, and Mamba models\cite{gu2023mamba}. The 3D CNN is responsible for capturing the implicit short-term temporal dynamics of eye movements, while the cascaded structure focuses on extracting explicit long-term temporal dynamics.

In the initial phase, the event-based eye tracking data is input into the 3D CNN, which effectively captures the fine-grained short-term temporal dynamics of the eye movements. The Frequency-Aware Module is then used to enhance the model's ability to focus on relevant frequency features that contribute to accurate eye movement tracking. Following this, the GRU and Mamba models are employed to analyze long-term temporal patterns, enabling the system to track eye movements across extended periods without losing accuracy.

Through the integration of these components, TDTracker achieves superior performance in eye movement prediction and localization, enabling real-time processing with minimal computational overhead. Additionally, a prediction heatmap is generated for precise eye coordinate regression, further improving tracking accuracy. This approach demonstrates state-of-the-art performance on event-based eye-tracking challenges, showcasing the effectiveness of combining temporal dynamics with advanced neural network architectures.

\paragraph{Implementation Details.}
Our server leverages the PyTorch deep learning framework and selects the AdamW optimizer with an initial learning rate set to $2\cdot e^{-3}$, which employs a Cosine decay strategy, accompanied by a weight decay parameter of $1\cdot e^{-4}$. This configuration is meticulously chosen to enhance the model's convergence and performance through adaptive learning rate adjustments. Training is conducted on an NVIDIA GeForce RTX 4090 GPU with 24GB of memory, enabling a batch size of 16. 

\paragraph{Results.}
\begin{figure}[t]
\centerline{\includegraphics[width= 7 cm]{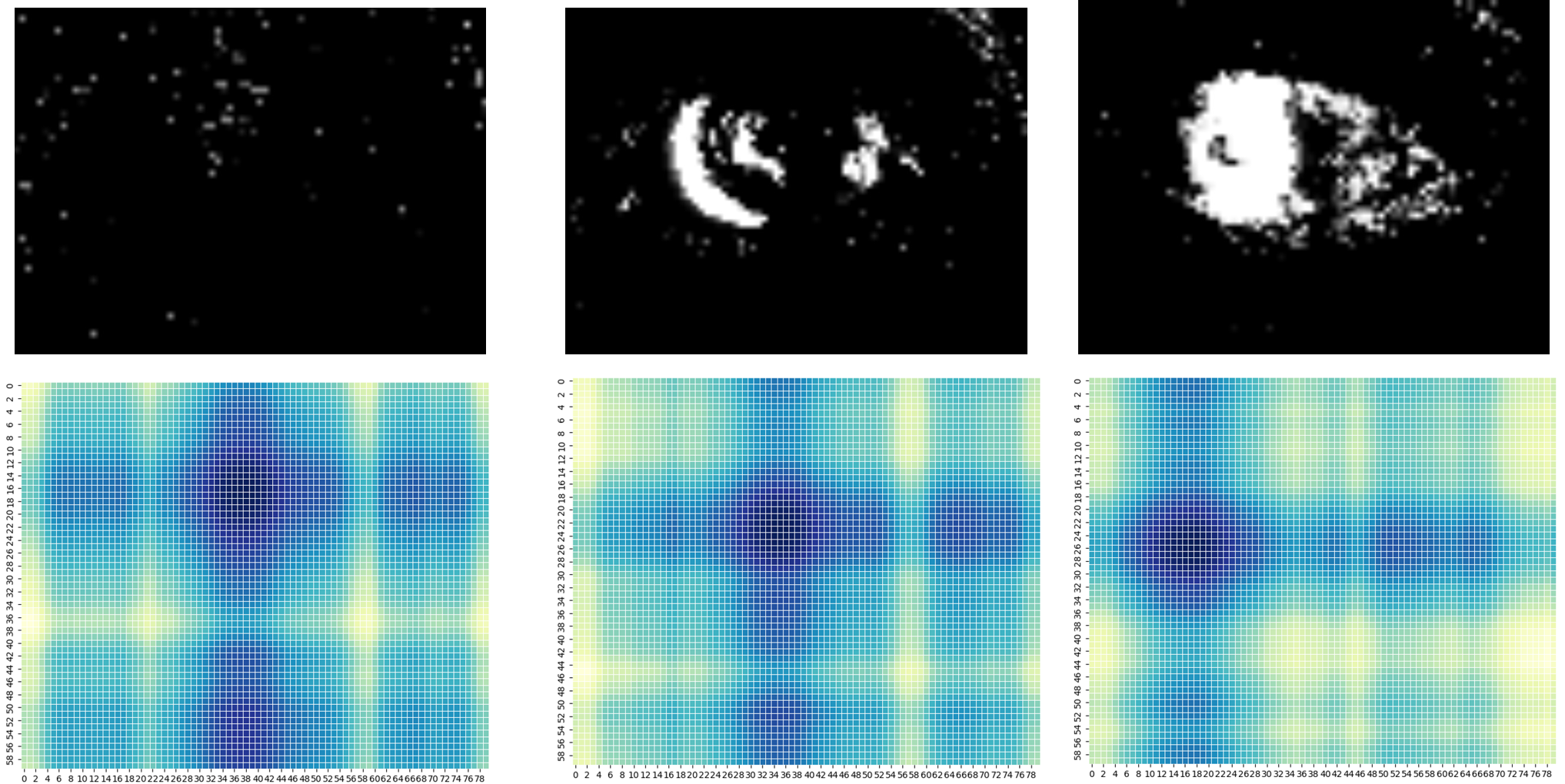}}
\caption{The visualization heatmap generated by the TDTracker.}
\label{fig: heat}
\end{figure}
\begin{table}[!t]
\centering
\small
\setlength{\tabcolsep}{5pt} 
\begin{tabular}{lccccc}
\hline
\textbf{Metric} & \textbf{Private} & \textbf{Public} & \textbf{Param} & \textbf{FLOPs} & \textbf{FPS} \\
\hline
TDTracker & 1.50861 & 1.46623 & 3.04M & 265M & 1.79\,ms \\
\hline
\end{tabular}
\caption{TDTracker’s performance on 3ET+ 2025.}
\label{tab:tdtracker_3et2025}
\end{table}
In the competition, we found that using 100 sequence training, 200 sequence testing had the highest accuracy (MSE: 1.62 to 1.55). However, since the parameters of the frequency-aware module are tied to the sequence length, we canceled this module during the competition. In addition, since our model does not consider open and closed eye cases, we simply use the ratio of the number of up and down events as the basis for judgment (set to 0.09), and when the current ratio is smaller than this value, the inference eye coordinate of the changed sample is overwritten by the inference value of the closest to this sample. What's more, we differ from directly regressing coordinate information by using a predicted probability density map, which provides an additional probability of the model predicting this image as shown in \cref{fig: heat}. If the probability is less than 0.5, we do not believe the predicted result. Tab.~\ref{tab:tdtracker_3et2025} shows the performance of TDTracker on the private and public test sets. After post-processing, the MSE is optimized to 1.4936 on the interpolation ground truth from 3ET+ 2024.

\subsection{Team: CherryChums}


\begin{center}

\noindent\emph{Hoang M. Truong, Vinh-Thuan Ly, Huy G. Tran,\\Thuan-Phat Nguyen, Tram T. Doan}

\noindent\emph{University of Science, Vietnam National University Ho Chi Minh City}

\noindent{\emph{Contact: \url{22280034@student.hcmus.edu.vn}}}

\end{center}

\begin{figure*}[h]
    \centering
    \includegraphics[width=0.7\linewidth, trim=0 15 0 10, clip]{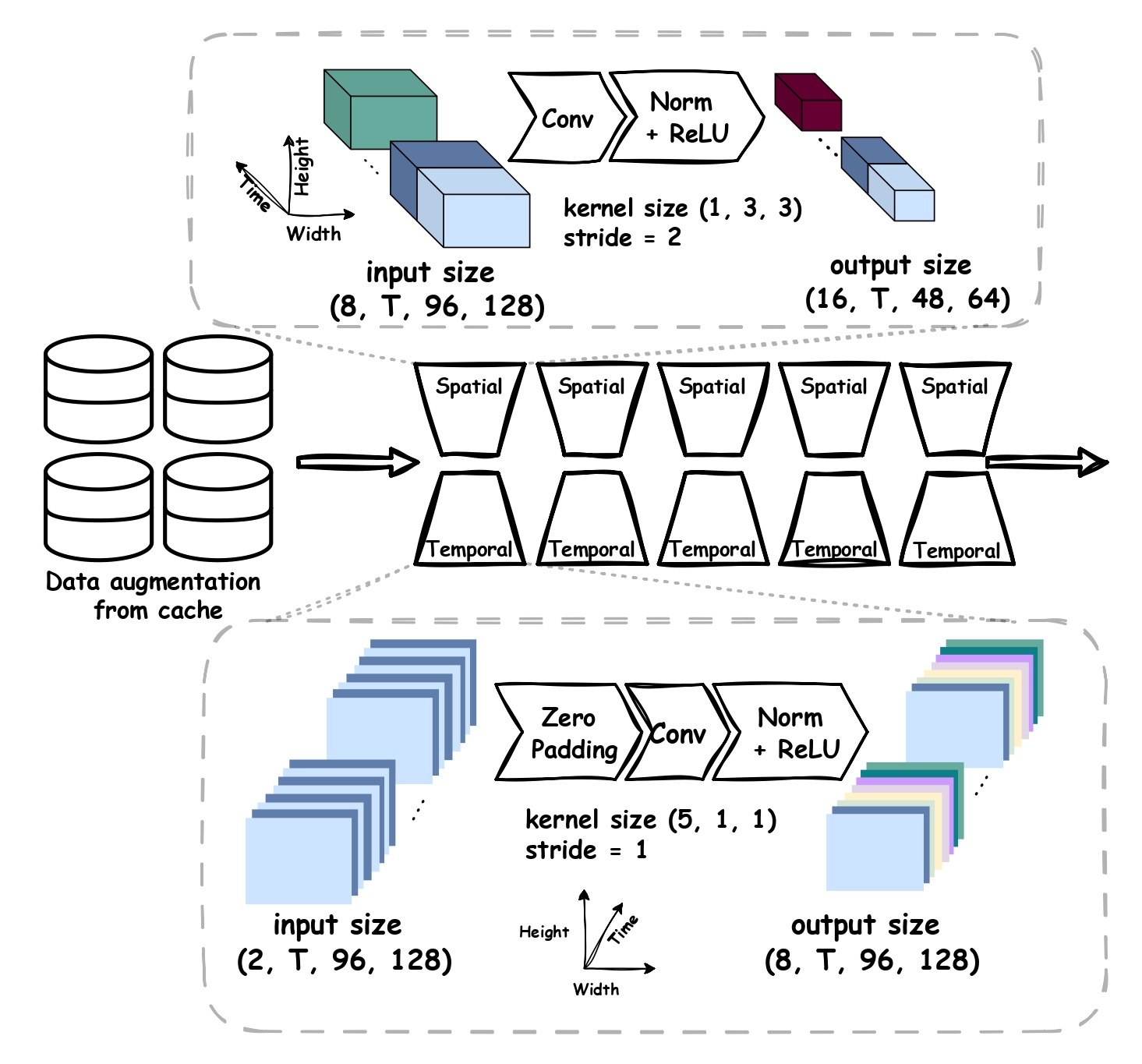}
    \caption{A compact spatiotemporal model integrating data augmentation with spatial and temporal processing blocks. Convolutional layers extract spatial and temporal features efficiently.}
    \label{fig:spatiotemporal_architecture}
\end{figure*}

\begin{figure}[h]
    \centering
    \includegraphics[width=1.0\linewidth, trim=30 20 26 23, clip]{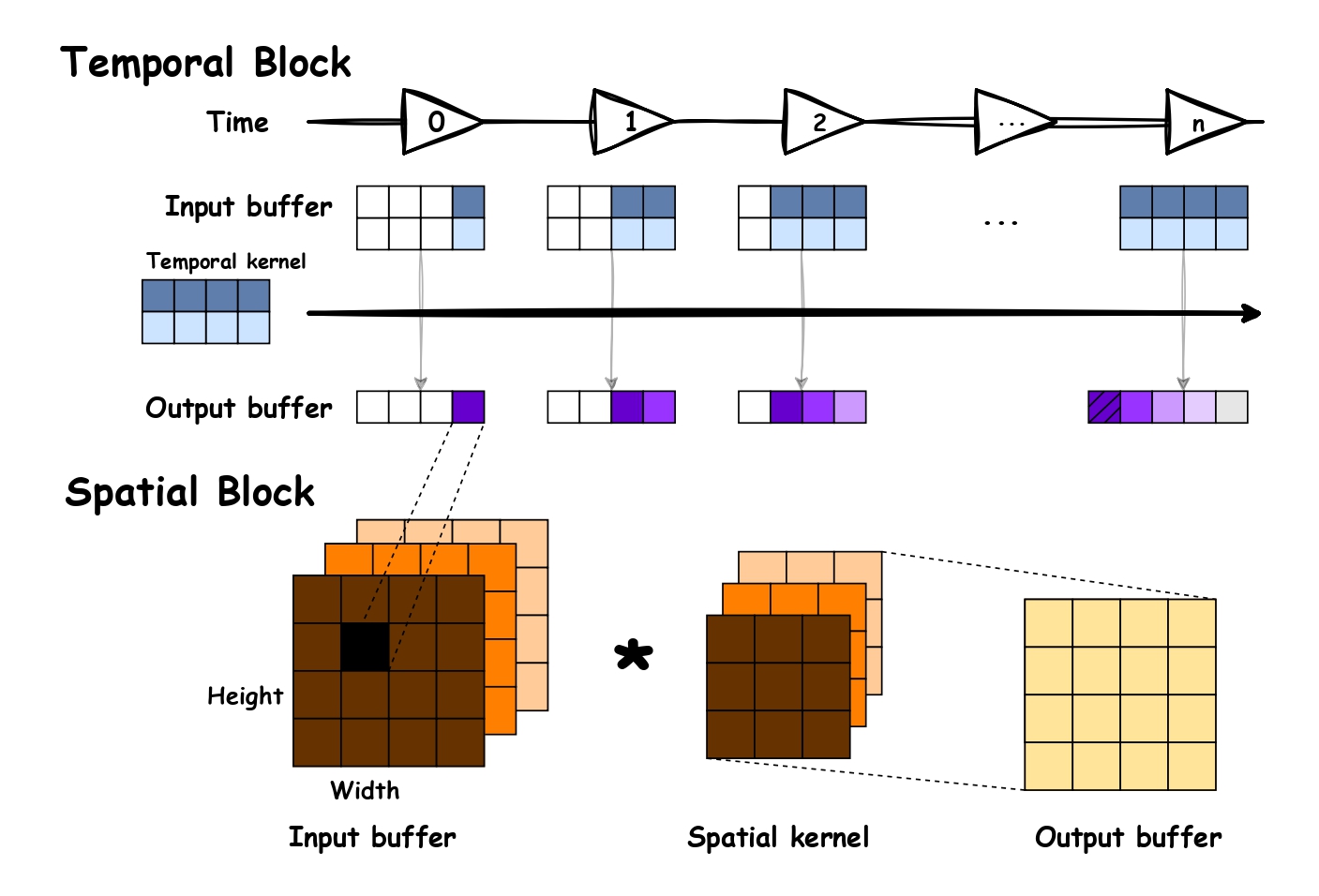}
    \caption{Illustration of spatiotemporal processing: the Temporal Block applies temporal convolution across frames, while the Spatial Block extracts spatial features using convolutional filters.}
    \label{fig:spatiotemporal_block}
\end{figure}

\paragraph{Description.}
The CherryChums team presents robust data augmentation strategies within a lightweight spatiotemporal network introduced by Pei \etal~\cite{pei2024spatiotemporal}. The network architecture is illustrated in~\cref{fig:spatiotemporal_architecture} and its spatiotemporal block in~\cref{fig:spatiotemporal_block}. This approach enhances model resilience against real-world perturbations, including abrupt eye movements and environmental noise.

\begin{figure}[t]
    \centering
    \includegraphics[width=1.0\linewidth, trim=21 0 23 0, clip]{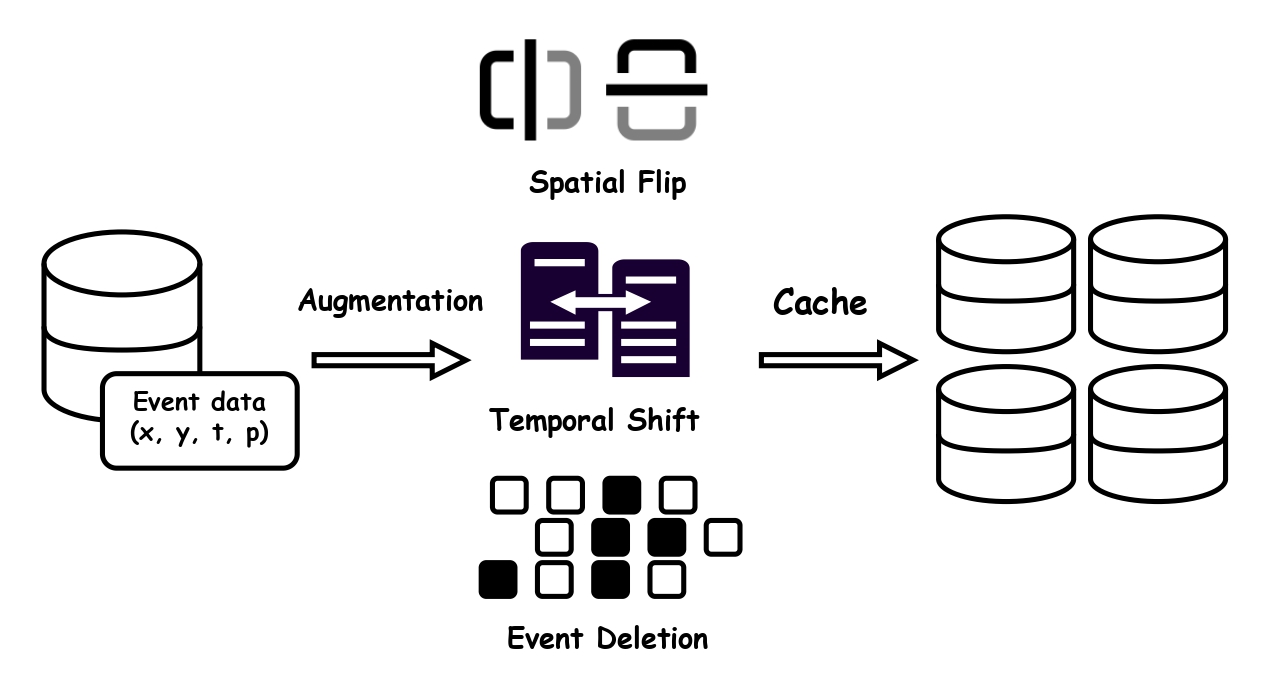}
    \caption{Overview of the data augmentation techniques used by CherryChums.}
    \vspace{10pt}
    \label{fig:augmentation_pipeline}
\end{figure}

The data augmentation pipeline, depicted in \cref{fig:augmentation_pipeline}, comprises temporal shift, spatial flip, and event deletion. These augmentations significantly bolster the model's robustness while preserving computational efficiency. More specifically:

\begin{itemize}
\item \textbf{Temporal Shift}: Given the asynchronous nature of event-based data, temporal augmentation is crucial for improving model resilience to timing variations. We apply a random shift to event timestamps within a range of $\pm200$ milliseconds while ensuring proper alignment of ground truth labels. Since labels are sampled at 100Hz (every 10ms), we recompute the label indices after shifting timestamps to maintain accurate correspondence.

\item \textbf{Spatial Flip}: To introduce spatial invariance, we horizontally and vertically flip the event coordinates $(x, y)$. The corresponding labels, including pupil center positions, undergo the same transformation to preserve spatial consistency.

\item \textbf{Event Deletion}: To simulate real-world sensor noise and occlusions, we randomly remove 5\% of the events while keeping the label sequence unchanged. This augmentation forces the model to learn from incomplete event streams and enhances its robustness to missing data.

\end{itemize}

The lightweight spatiotemporal network is optimized for real-time performance on edge devices, leveraging causal spatiotemporal convolutional blocks and efficient event binning. By combining these techniques, our method achieves improved accuracy and robustness in event-based eye tracking.

\paragraph{Implementation Details.}
For the lightweight spatiotemporal network, we adopt the original training configuration from Pei \etal~\cite{pei2024spatiotemporal}. Specifically, we train the model using a batch size of 32, where each batch contains 50 event frames. The network is optimized for 200 epochs using the AdamW optimizer with a base learning rate of 0.002 and a weight decay of 0.005. We employ a cosine decay learning rate schedule with linear warmup, where the warmup phase spans 2.5\% of the total training steps. Additionally, we leverage automatic mixed-precision (AMP, FP16) and PyTorch compilation to accelerate training and improve efficiency. All experiments were conducted on a single NVIDIA Tesla P100 GPU provided by Kaggle.

\paragraph{Results.}
Our approach demonstrated significant improvements in robustness and accuracy, achieving a Euclidean distance error of 1.61, compared to 1.70 by the original spatiotemporal network trained without our additional augmentation strategy, as shown in \cref{tab:compare_metrics}. The results highlight the effectiveness of our data augmentation strategy in enhancing model performance under challenging conditions.

\begin{table}[ht]
    \centering
    \begin{tabular}{cccc}
        \toprule
        \textbf{Year} & \textbf{Augmentation} & \textbf{pixel error} & \textbf{p\_10} \\
        \midrule
        \multirow{2}{*}{2024}  
        & \ding{55}  & --     & 99.16 \\
        & \ding{51} & --     & \textbf{99.37} \\
        \midrule
        \multirow{2}{*}{2025}
        & \ding{55}  & 1.70  & -- \\
        & \ding{51} & \textbf{1.61}  & -- \\
        \bottomrule
    \end{tabular}
    \caption{Comparison of Euclidean Distance and p\_10 Accuracy in both 2024 and 2025 event-based eye tracking challenges.}
    \label{tab:compare_metrics}
\end{table}

As part of our challenge report, we provide a summary of the model size and the number of MACs required per frame, as shown in~\cref{tab:model_summary}. This information is crucial for understanding the efficiency and real-time deployment capabilities of our network.

\begin{table}[ht]
\centering
\begin{tabular}{lc}
\toprule
\textbf{Metric}       & \textbf{Value} \\
\midrule
Parameters            & 807~K         \\
MACs per Frame        & 55.18~M       \\
\bottomrule
\end{tabular}
\caption{Summary of Model Size and MACs}
\label{tab:model_summary}
\end{table}


\section{\textcolor{black}{Hardware Discussion}}
\label{sec:discussion}

The hardware design of event-based eye tracking systems necessitates careful consideration of power efficiency, latency, and adaptability to ensure robust performance in real-world applications. 
A central advantage of event-based architectures lies in DVS’s inherent power efficiency, which stems from its sparse data generation. Unlike conventional frame-based systems that continuously sample and transmit data, event-based systems respond only to changes in retinal activity, drastically reducing redundant data throughput. 
Specialized event-driven hardware designs that leverage this unique property of DVS have the potential to achieve high-performance tracking during rapid eye movements (saccades) while minimizing redundant processing during fixation periods. This efficiency can be enabled through optimization strategies such as event-triggered circuit architectures, which dynamically activate computational resources only in response to detected motion, thereby aligning power consumption with real-time demands.
Additionally, this approach aligns with emerging paradigms in near-sensor or in-sensor processing, where computation is localized near the sensing element to minimize data transmission to external processors. Recent studies have showed that in-sensor event filtering and preprocessing could reduce communication bandwidth, thereby lowering both latency and power consumption~\cite{feng2024blisscam}.

A critical challenge in event-driven hardware design involves balancing the relationship between processing latency and sensor’s sampling rate. While DVS's high sampling rate theoretically improves temporal resolution, it also brings a potential risk of overwhelming downstream processing pipelines, leading to data congestion or potential data leakage. Therefore, optimizing buffering strategies and parallel processing architectures becomes important to handle sporadic bursts of events without introducing bottlenecks. 
For instance, integrating dedicated memory hierarchies or distributed processing units can prevent data contention during high-activity intervals. 
Such optimizations ensure that the hardware maintains low-latency responsiveness, critical for applications like foveated rendering \cite{liu2025fovealnet} and real-time human-computer interaction, while avoiding unnecessary computational overhead during periods of eye fixation.

Moreover, configurability emerges as a pivotal design consideration to enhance the adaptability of event-based eye tracking systems across diverse operational environments. Preferably, hardware should have support for tunable parameters, such as event detection thresholds, temporal filtering windows, or region-of-interest (ROI) prioritization, to accommodate variations in lighting conditions, user ergonomics, or application-specific requirements. 
For example, dynamic reconfiguration of event thresholds could optimize sensitivity in low-light environments, while selective ROI processing could conserve resources by focusing computation on critical areas of the visual field. Hardware designs with this flexibility can not only extend the system’s utility across different use cases but also future-proof the design against evolving algorithmic demands. 
By embedding configurability into the hardware architecture, designers can strike a balance between generality and efficiency, ensuring that event-based eye tracking systems remain both practical and scalable in real-world deployments.

{\small
\bibliographystyle{ieeenat_fullname}
\bibliography{shared_bib, 
bibs/USTCEventGroup}
}

\end{document}